\documentclass[conference]{IEEEtran}
\usepackage[T1]{fontenc}
\usepackage[table]{xcolor}
\usepackage{booktabs}
\usepackage{array}
\usepackage{amsmath}
\usepackage{amssymb}
\usepackage{graphicx}
\usepackage{tabularx}
\usepackage{balance}
\usepackage[hidelinks]{hyperref}
\hypersetup{pdfauthor={}, pdftitle={EdgeLLM-Nano}}
\title{\textbf{EdgeLLM-Nano:}\\
Auditing Polar-Guided Key-Cache Compression and Consistency-Anchored Pruning\\
on a 33.6\,MB On-Device Language Model}
\author{
\IEEEauthorblockN{
S M Mahbub Murshed\IEEEauthorrefmark{1},
Hayden Bright\IEEEauthorrefmark{2},
Rezaul Karim\IEEEauthorrefmark{3},
Ali Shiri Sichani\IEEEauthorrefmark{2}
}
\IEEEauthorblockA{
\IEEEauthorrefmark{1}Causal Dynamics Lab
}
\IEEEauthorblockA{
\IEEEauthorrefmark{2}Electrical Engineering and Computer Science Department,\\
University of Missouri-Columbia
}
\IEEEauthorblockA{
\IEEEauthorrefmark{3}Computer Science \& Engineering, University of Chittagong
}
\IEEEauthorblockA{
murshed@gmail.com, haydenabright@gmail.com,\\
rezaul.cse@cu.ac.bd, asp9f@missouri.edu
}
}
\begin{document}
\maketitle

\begin{abstract}
Deploying language models (LMs) on the edge
presents unique memory, latency, and power challenges absent in
high-performance computing. KV-cache compression emerges as
a promising approach to address these challenges. However, the
effectiveness of current KV-cache compression methods has not
been thoroughly examined on edge devices. Moreover, we show
that existing evaluation methods do not always involve the KV-cache
 under test. We propose EdgeLLM-Nano, a measurement
study of KV-cache compression using a nano-scale decoder-only
 LM (transformer) testbed that has 8.39M parameters, i.e.,
33.6 MB, to enable end-to-end examination of each backend,
seed, and power sample. Our results give a negative verdict
for a polar-guided key-cache quantizer (round-to-nearest (RTN)
control causes an order of magnitude less damage), as well
as a consistency-anchored progressive weight pruner (one-shot
magnitude-based pruning performs as well as, if not better). We
also introduce Long-Context Decode Fidelity (LCDF) evaluation,
a token-by-token examination of teacher-forced log-losses using
the KV-cache under test. Unlike perplexity that gives the same
5.461 score to all methods, LCDF distinguishes different KV-cache
 compression backends thanks to its paired statistics across
seeds and effect sizes. Our companion Raspberry Pi 5 measurement
 shows all compression backends we tested to be less energy-efficient
 per token compared to uncompressed FP16 inference.
Except for KIVI that is close to lossless in terms of LCDF,
our findings indicate KV-cache compression can be improved.
LCDF offers a simple and effective way of shedding light on
KV-cache compression methods, whose negative impacts have
been overlooked by perplexity.
\end{abstract}


\section{Introduction}
\label{sec:intro}
Single-shot perplexity is inadequate for evaluating KV-cache compression because it does not write or read through a compressed cache. From the experiments every backend reports the same perplexity (5.461), from lossless FP16 through aggressive quantization. In addition, KV-cache compression requires persistent correctness throughout autoregressive decoding (not merely during calibration). To address these gaps, we make the following contributions:

EdgeLLM-Nano is introduced as a compact end-to-end testbed for KV-cache compression, targeting realistic on-device inference rather than aggregate FLOPs alone. Using an 8.39\,M parameter decoder-only LLM (33.6\,MB weights; trained on TinyStories) running on a 4\,GB Raspberry Pi~5 SBC, we measure accuracy, speedup, cache size, and per-token latency under representative workloads. At context 512, the model's FP16 KV-cache is at most 3.1\,MB on a 4\,GB SBC: compression may improve efficiency, but is not required to make this workload feasible. We include a second, larger repetition using a 124-million-parameter LLM showing that numerical rankings do not extrapolate beyond the measured regime.

Long-Context Decode Fidelity (LCDF) is proposed as a perplexity-style, token-level teacher-forced evaluation that exposes compression-induced fidelity loss invisible to single-shot perplexity, with paired, seed-level statistics and effect sizes. Together with the testbed, LCDF enables more faithful evaluation of KV-cache compression for LLMs.

Polar-Guided Key-Cache Quantization (PGKQ) with negative result is proposed. PGKQ applies a Hadamard transform, a polar decomposition, a Lloyd--Max quantizer, and a 1-bit Quantized JL residual (via a sign-projection trick), reaching 3.75 bits per element for the key-side cache. Our findings: PGKQ's rotation and polar steps are independently established \cite{wu2025polarquant, su2025rotatekv}, PGKQ underperforms a simple, matched-bits round-to-nearest quantization baseline for K-side cache quantization, and the widely-used KIVI quantizer \cite{liu2024kivi} achieves superior results to PGKQ for both the K- and V-side caches and is nearly lossless.

Negative results for Progressive Self-Supervised Pruning with Consistency Anchoring (PSPCA) for LLM compression is reported. It applies a Wanda-scored pruning schedule shaped cubically over pruning rounds followed by 200 recovery fine-tuning (i.e., AdamW) steps, i.e., PSPCA is not retraining-free \cite{sun2024wanda, frantar2023sparsegpt}. PSPCA is not found to be superior to simple one-shot magnitude-based pruning.

Context-Sensitive Energy-Budget Routing remains an unevaluated design study rather than a claimed contribution \cite{shazeer2017moe, schuster2022calm}.

\section{Related Work}
\label{sec:related-work}

\begin{table*}[!t]\centering\footnotesize\setlength{\tabcolsep}{3pt}
\caption{Method positioning and LCDF verdicts on this testbed}
\label{tab:positioning}
\begin{tabularx}{\textwidth}{@{}l l l X c c X@{}}
\toprule
Method & Venue & Lane & Core mechanism & Bits/elt & Cache eval & LCDF verdict \\
\midrule
KIVI \cite{liu2024kivi} & ICML'24 & K$+$V & Per-channel K, per-token V, tuning-free & 2.99 / 3.79$^\dagger$ & Yes & Near-free: $+0.0053$; $+0.0059^\dagger$ \\
KVQuant \cite{hooper2024kvquant}$^\ddagger$ & NeurIPS'24 & K$+$V & Calib.\ per-channel $+$ outlier fallback & 3.50 & Yes & Small, significant: $+0.0254$ \\
QuaRot \cite{ashkboos2024quarot} & NeurIPS'24 & K$+$V & Global Hadamard $+$ RTN & 4.00 & Yes & Small, significant: $+0.0117$ \\
SnapKV \cite{li2024snapkv} & NeurIPS'24 & K$+$V & Attention-guided eviction & 16, no eviction & Yes & Jitter below $10^{-4}$; wrapper overhead \\
QJL \cite{zandieh2025qjl} & AAAI'25 & K & 1-bit JL sign projection & PGKQ component & Yes & Never earns its $+1.25$ bits here \\
RotateKV \cite{su2025rotatekv} & IJCAI'25 & K$+$V & Channel reorder $+$ FWHT & --- & Yes & Not measured; shares PGKQ steps 1--2 \\
PolarQuant \cite{wu2025polarquant} & arXiv'25 & K & Polar transform of keys & --- & Yes & Not measured; shares PGKQ step 3 \\
\rowcolor{gray!12} RTN-INT3 control & this paper & K & Per-token asym.\ round-to-nearest & 3.50 & LCDF & $+0.025$/$+0.006$; beats PGKQ 11--20$\times$ \\
\rowcolor{gray!12} PGKQ, this work & this paper & K & Reorder$+$FWHT$+$polar$+$Lloyd--Max$+$QJL & 3.75 & LCDF & $+0.19$ mean; dominated by RTN \& KIVI \\
\bottomrule
\end{tabularx}
\par\vspace{2pt}\parbox{\textwidth}{\footnotesize Bits per element are audited: baseline rates are nominal, whereas PGKQ includes metadata. LCDF $\Delta$ is the mean against FP16 over six datasets at ctx$=512$ (Table~\ref{tab:lcdf}). $^\dagger$Reference streaming policy with three prompts. $^\ddagger$K-side port diverges from the reference. RTN uses two datasets.}
\end{table*}

The two proposed compression mechanisms, four ported baselines, and a matched-bits control under one decode-through-cache protocol are presented in the Table~\ref{tab:positioning}.

\textit{Quantization:} RotateKV \cite{su2025rotatekv} couples outlier-aware channel reordering with a fast Walsh--Hadamard transform (FWHT), and PolarQuant \cite{wu2025polarquant} quantizes keys in polar coordinates; both correspond to the first three steps of PGKQ. PGKQ additionally applies a Lloyd--Max radius codebook and a QJL sign residual \cite{zandieh2025qjl}. From our experiments, this combination does not outperform round-to-nearest quantization at matched bits (Section~\ref{sec:results}) and include PGKQ to make this negative result auditable. Other work targets complementary axes we do not measure: adaptive, per-layer budgets; low-rank components; and query-agnostic eviction \cite{shutova2025aquakv, chang2025palu, li2024snapkv, kim2025kvzip}.

\textit{Evaluation methodology:} The single-shot perplexity such as the present pipeline and several similar-scale studies were evaluated. We do not claim this is true of the literature, and note that this protocol does not exercise the compressed cache. KV cache compression studies report generation-from-cache on long-context suites (e.g., LongBench, needle-in-a-haystack tests, GSM8K) \cite{bai2024longbench}. We agree with recent calls for more consistent evaluation \cite{li2025kvsurvey, javidnia2025keyvaluecompress, gao2025rethinkingkv} and contribute matched seed--prompt teacher-forced log-loss, a cache-breaking protocol to validate implementation correctness, and cost-effective reproducibility at this model scale.

\textit{Pruning:} Magnitude pruning with subsequent fine-tuning for performance recovery dates at least to Han et al.\ \cite{han2015learning}. Subsequent work has enabled pruning without fine-tuning (i.e., zero gradient steps): SparseGPT and Wanda perform one-shot pruning \cite{frantar2023sparsegpt, sun2024wanda}. Structured N\,:\,M pruning has received renewed interest for compatibility with sparse tensor cores \cite{mishra2021nm}. The closest prior work to PSPCA uses the same importance score as Wanda within a cubic pruning schedule. Unlike Wanda and SparseGPT, PSPCA is not retraining-free because it performs recovery via anchor-based fine-tuning. Our results suggest one-shot magnitude pruning performs comparably or better to PSPCA at matched sparsity (Section~\ref{sec:results}).

\section{Method}
\label{sec:method}

\subsection{NanoGPT-256 base architecture and LCDF}
\label{sec:nanogpt-arch}
\textit{Base model:} The base model is a decoder-only NanoGPT-256 Transformer with 6 layers, 4 heads, model width of 256, FFN width of 1\,024 and 512-token context. We use RMSNorm \cite{zhang2019rmsnorm}, rotary position embeddings \cite{su2024roformer}, a SwiGLU FFN \cite{shazeer2020glu} and a weight-tied embedding and output head. We train a byte-level (char) model, as well as a SentencePiece BPE model (Table~\ref{tab:base-training} reports the latter).

\textit{LCDF evaluation:}\label{sec:lcdf} We use the Long-Context Decode Fidelity (LCDF) introduced above. In detail, given a shared prompt, our evaluation harness decodes the compressed cache one-token-at-a-time and reports teacher-forced next-token log-loss at each decode position. Each reported LCDF is the delta with respect to decoding with an FP16 cache for the same checkpoint and prompt, with identical decoded tokens. All reported LCDFs are thus paired at the seed--prompt level to control for the quality of the base LM and isolate LM degradation due to cache compression. For statistical significance, we report Cliff's $\delta$ as an effect size metric. To verify that our harness is able to catch destructive KV cache modification, we perform a unit-test with a purposefully broken KV cache which results in large LCDF increase as desired.

\subsection{PGKQ --- Polar-Guided Key-Cache Quantization}
\label{sec:pgkq}
\begin{figure}[t]
\centering
\includegraphics[width=\columnwidth]{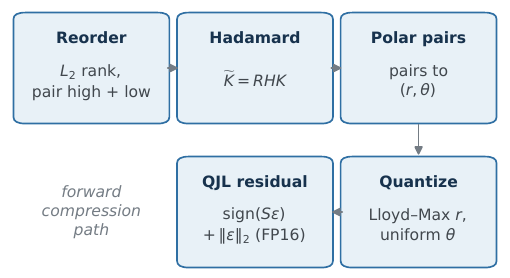}
\caption{PGKQ's forward compression path. The first four stages form the main polar code. The optional QJL stage stores a sign-projected residual and its norm.}
\label{fig:pgkq-pipeline}
\end{figure}

\textit{PGKQ pipeline:} Given calibration keys $K \in \mathbb{R}^{N \times d}$: (0) optionally reorder channels in decreasing order of $\ell_2$ norm and group them with the goal of putting one high-rank and one low-rank channel in the same group; (1) precondition $K$ with a randomized Hadamard transform (HT): $\tilde K = RHK$; (2) group adjacent components in the HT-domain and convert them into a pair of polar coordinates, a radius $r_i$ and angle $\theta_i$; (3) quantize the radii with a $b_{\text{main}}$-bit scalar codebook and angles on a $b_{\text{angle}} = b_{\text{main}}-1$-bit uniform grid on $[-\pi,\pi]$ to obtain $\hat r$ and $\hat\theta$; and (4) optionally compute residual $\varepsilon = K-\hat K$ and store a Quantized Johnson--Lindenstrauss (QJL) representation, a projection onto $m=\text{head\_dim}$ binary random directions $\text{sign}(S\varepsilon) \in \{\pm1\}^m$, plus a single FP16 number, the residual norm $\|\varepsilon\|_2$ per token. At inversion time, we never explicitly invert the codes to construct $\hat K$. Instead, attention logits are computed analytically given $Q$, $\hat r$, $\hat\theta$ and $\hat\varepsilon$ when QJL is used. Figure~\ref{fig:pgkq-pipeline} summarizes PGKQ.

For the radii, we use the codebook obtained by Lloyd--Max for the Beta prior of \cite{lloyd1982quantization, max1960quantizing}. For simplicity, we do not refit the codebook empirically and use this codebook for all results in Table~\ref{tab:kv-comparison-full} and Figure~\ref{fig:pi5-energy}. For example, at $b_{\text{main}}=3$ and $b_{\text{angle}}=2$, using QJL with a residual projection to 64 dimensions, PGKQ uses a total of 485\,376 bytes for the K$+$V cache for a 128-token cache as opposed to 786\,432 bytes for FP16. Thus, even without compressing $V$, PGKQ leads to a $1.62\times$ reduction in K$+$V cache memory footprint in this example. This agrees with the 3.75 bits/element for keys that we report in the introduction.

\subsection{PSPCA --- Progressive Self-Supervised Pruning with Consistency Anchoring}
\label{sec:pspca}
Specifically, PSPCA consists of the following two phases. In the first phase, the output logits on the calibration data as well as the outputs of all Transformer blocks are saved as anchors. Notably, PSPCA does not require task-specific labels during this phase.

In the second phase, the pruning sparsity $s_k$ progressively increases over a total of $n_{\text{rounds}}=8$ pruning and recovery rounds, with
\begin{equation}
s_k = s_{\text{target}}\left(\frac{k+1}{n_{\text{rounds}}}\right)^{3},
\label{eq:pspca-schedule}
\end{equation}
where $k = 0,\dots,7$ denotes the pruning and recovery round index. At the beginning of each round, we compute the importance score of each parameter using the same method as in Wanda \cite{sun2024wanda}: $I(w) = |W|\cdot\|x\|$, where $W$ is the weight matrix and $x$ are activations recorded during calibration. Then we perform pruning according to the computed importance scores for all linear layers except the embedding layer and the task-specific output heads. After pruning, we perform a short fine-tuning phase for $n_{\text{rec}}=200$ gradient steps using AdamW optimizer and a matching objective function, i.e., minimizing the $L_2$ distance between the hidden states and the anchors, considering all Transformer blocks.

\section{Experimental setup}
\label{sec:setup}

\textit{Datasets:} One BPE checkpoint $\times$ eight KV-cache backends are tested on six datasets (TinyStories \cite{eldan2023tinystories} (ID); SQuAD~v2, PubMedQA, DailyDialog (OOD); ClinTemplate-synth, IoT-NLP-synth (synthetic OOD)) with $n=10$ random seeds; each seed contributes 30 runs. TinyStories uses the full training split (2.12\,M rows). SQuAD~v2 uses the standard Hugging Face validation split. PubMedQA uses the \texttt{pqa\_labeled} subset of PubMedQA, taking the title and abstract as the context and the question as the prompt, with answers not used by the compressor. DailyDialog is a community-mirrored Hugging Face dataset provided as parquet files (Hugging Face parquet mirror) and may differ from the canonical release. IoT-NLP-synth uses templated prompts (by design). ClinTemplate-synth's MedQA portion is unavailable: the loader is deprecated, and all rows fell back to the same templated default (see the dataset card). Its manifest still lists \texttt{medqa}. Protocol perplexity (higher is harder) is roughly 200 to 600 on four datasets but 12.6 on DailyDialog; we therefore focus the main claims on the in-domain set (TinyStories), report all six for transparency, and flag DailyDialog as an easy-format outlier.

\textit{Implementation details:} All models are trained with the AdamW optimizer \cite{loshchilov2019adamw} and a cosine LR schedule decaying to zero using bf16 mixed-precision training. The NanoGPT-256-BPE (Table~\ref{tab:base-training}) trains for 20\,000 steps using a batch size of 32 sequences of length 512 for approximately 10 minutes using SentencePiece BPE Tokenizer trained on TinyStories dataset with a vocab size of 8\,192 tokens. The GPT-2-124M is trained for 2\,000 steps using a batch size of 8 sequences of length 1\,024 for approximately 8.5 minutes on TinyStories dataset with the original GPT-2 Tokenizer. We used greedy decoding. PGKQ settings use 64-dim residual projections. Table~\ref{tab:kv-comparison-full} and Figure~\ref{fig:pi5-energy} fit the codebook using random IDs as Beta Prior with shape of $8\times4\times512$ (batch $\times$ seq $\times$ head) using Lloyd--Max Quantizer, Table~\ref{tab:lcdf} uses empirical fit with captured activations, which improves (halves) the quality-cost (Figure~\ref{fig:gate-m1-pareto}). PSPCA uses 2\,048 anchor/calibration token IDs. Channel reordering is enabled unless noted.

\textit{Baselines:} All baselines are implemented by us as-is (no tuning), with the following exceptions. KIVI uses the authors' recommended defaults of 2/4-bit K/V \cite{liu2024kivi}. Our local KVQuant-like baseline (not the authors' code) uses 3/4-bit K/V with FP16 residual outliers (default fraction 1\%) \cite{hooper2024kvquant}. QuaRot uses 4-bit K/V \cite{ashkboos2024quarot}. SnapKV's recommended eviction budget equals our maximum tested context length. Therefore, it evicts no tokens here (thus measuring wrapper/overhead effects rather than compression, and it has no bits-per-element axis) \cite{li2024snapkv}.

\textit{Validation:} The KIVI port reproduces the reference's quantized integers; dequantized values agree to a max abs diff of $10^{-6}$ and LCDF logits to a max abs diff of $1.2\times10^{-7}$ \cite{liu2024kivi}. KIVI's default residual-windowed group processing is streaming-oriented and does not engage in our token-by-token harness; omitting it makes the K-lane effectively lossless. Under the authors' exact settings (k2v2, g32, r32), a quick 3-prompt check shows delta is $+0.0059$ at 3.79 bpe; conclusions are unchanged. Our local KVQuant-style baseline reproduces the authors' V-lane quantizer indices on all 32\,768 elements for a random uniform kernel, but its K-lane differs materially: we quantize post-RoPE (with affine scaling and residual outliers), whereas KVQuant calibrates and quantizes pre-RoPE with a learned, typically nonuniform mapping; outlier-mask agreement is accordingly poor (e.g., $<1\%$).

\textit{Accounting:} Baseline KV-byte counts exclude per-block scales/outlier metadata; ours include it. This favors baselines.

\textit{Statistics:} We report 95\% CIs using 10,000 bootstrap replicates ($B=10{,}000$) on mean per-seed values for throughput, p99, KV-bytes and ppl. LCDF (lower is better) uses a Wilcoxon signed-rank test versus FP16 (two-sided). Pareto uses a Wilcoxon signed-rank test (two-sided) per point on the swept curve against the corresponding KIVI point using paired sweeps with the same 30 sequences.

\textit{LCDF details:} For each dataset, we compute a mean-delta per random seed over three prompts (10 values per backend), then Wilcoxon-test these 10 values versus FP16 to avoid treating correlated samples as independent (which would understate p-values). We apply Holm--Bonferroni across the seven backends within each dataset. The primary axis is ctx$=512$, with ctx$=128/256$ reported for transparency. Of 84 method $\times$ dataset comparisons at 128/256, 82 agree in sign with the primary; both exceptions have absolute delta $<5\times10^{-4}$. All experiments were conducted on a single NVIDIA RTX~5090 GPU (32\,GB RAM) with PyTorch 2.12 (nightly) and CUDA 13.0.2. Edge results are obtained using Raspberry Pi~5 (4\,GB RAM), INA219 Power Sensor and Linux 6.12.75 64-bit OS. The minimum two-sided p-value we observe is roughly 0.002.

\section{Results}
\label{sec:results}

\subsection{Base-model training and PGKQ K-side compression}
\label{sec:pgkq-bits-results}
\begin{table}[t]\centering\footnotesize\setlength{\tabcolsep}{3pt}
\caption{Base-model training summary}
\label{tab:base-training}
\begin{tabular}{l c c c c c}
\toprule
Model & Tokenizer & Vocab & Params & Steps & Val ppl \\
\midrule
NanoGPT-256 char & byte & 256 & 6.36\,M & 20\,000 & 1.846 \\
NanoGPT-256 BPE & SP-BPE & 8\,192 & 8.39\,M$^*$ & 20\,000 & \textbf{5.117} \\
GPT-2-124M scale & tiktoken-gpt2 & 50\,257 & $\sim$124\,M & 2\,000 & 8.225 \\
\bottomrule
\end{tabular}
\par\vspace{2pt}\parbox{\columnwidth}{\footnotesize Parameter counts include the tied embedding; $^*$6.29\,M without it. Val ppl is the mean of 16 batches at bs$=32$. Section~\ref{sec:setup} gives the dtype and wall-clock time.}
\end{table}

\textit{Bit budget:} PGKQ in the text is evaluated with $b_{\text{main}}=3$, $b_{\text{angle}}=2$, and qjl\_m$=64$. The number of bits per element required to store these parameters is $(b_{\text{main}}+b_{\text{angle}})/2$ per element to store polar coords for each 2-element block, 1 bit per element for QJL signs, and a total of $16/\text{qjl\_m}$ per element to store the residual norms in FP16 for every qjl\_m elements within each head. The final audited number for PGKQ is 3.75 bits per element on the K side (and PGKQ does not apply to V), which depends on the model's transformer head\_dim as well as $b_{\text{main}}$, $b_{\text{angle}}$, and qjl\_m but not vocab size. We report PGKQ K-only results in Figure~\ref{fig:gate-m1-pareto} as 9.86 bits per element by averaging FP16 (16 bits per element) V with PGKQ 3.75 bit K for ctx$=512$ but accounting for 511 occupied slots in the KV-cache.

\textit{Ablations for PGKQ:} For each of TinyStories and SQuAD-v2, we conduct a $2\times2$ experiment to understand the effect of each of Hadamard rotation and QJL residuals, and compare LCDF $\Delta$ at ctx$=512$ as follows. Each experiment was conducted on 30 samples using CPU-side seed--prompt inference as in Section~\ref{sec:setup}, with the same prompts and protocol. The empirical codebook was computed with $b_{\text{main}}=3$ as in PGKQ above, leading to a total bit budget of 3.75 bits per element with QJL (as above), and 2.50 without QJL. On TinyStories, $\Delta$ is 0.282 as reported, 0.483 without Hadamard or QJL, 0.265 with only Hadamard, and 0.250 (p-value of 0.01) with only QJL, suggesting that each contributes but that QJL residuals are not worth their bit budget (1.25 bits per element). On SQuAD-v2, however, we find that using only Hadamard with 2.50 bits per element actually performs best, reducing $\Delta$ from 0.128 to 0.036 (p-value of 0.002), with the full method performing the worst. These results suggest that the effects of each of these techniques may not be additive, and may be dataset/task dependent.

\textit{Comparison to simpler baselines:} Given the surprising results above, we revisit the need for PGKQ's polar-coord transformation, Lloyd--Max codebooks, Hadamard rotation, and QJL residuals by comparing against per-token asymmetric round-to-nearest (RTN) quantization of K with a matched bits budget, keeping V in FP16, with the same seed--prompt protocol above. RTN requires storing two params (scale, zero-point) in FP16 for each quantized vector, resulting in a total bits-per-element budget of 3.50 or 4.50 for INT3 or INT4 quantization of K (to account for the extra 0.5 bits per element needed to store the two params in FP16). On TinyStories, this improves $\Delta$ over PGKQ, reducing it significantly (p-value of 0.002) to 0.025 from 0.282 (PGKQ) for INT3 (fewer bits per element!), and further reducing it to 0.008 for INT4. On SQuAD-v2, RTN also performs better than PGKQ, significantly (p-value of 0.02) improving $\Delta$ to 0.006 with INT3 from PGKQ's 0.128, and further (not significantly) to 0.002 for INT4. These results suggest that the tested version of PGKQ underperforms a simpler RTN baseline. We hesitate to conclude more than that the tested setup underperforms RTN baselines, leaving it to future work to revisit PGKQ design decisions.

\subsection{PGKQ-KV: extending compression to the V-side}
\label{sec:pgkq-kv}
\begin{figure}[t]
\centering
\includegraphics[width=\columnwidth]{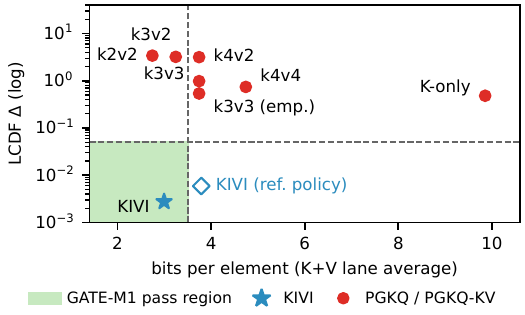}
\caption{GATE-M1 Pareto sweep over 30 TinyStories sequences at ctx$=512$. This single-dataset result is not comparable to Table~\ref{tab:lcdf}. The axes show lane-averaged cache-hook bits per element and log-scale LCDF $\Delta$. KIVI lies inside the pass region; the filled star uses our port and the open diamond uses the reference policy. Every PGKQ-KV variant lies outside. The K-only point at 9.86 bits and 0.480 averages the 3.75-bit K lane with FP16 V.}
\label{fig:gate-m1-pareto}
\end{figure}

PGKQ-KV additionally applies PGKQ's polar representation, Lloyd--Max codebooks and QJL projection to V; like PGKQ it may use PGKQ's Beta-prior codebooks (PGKQ-KV) or a PGKQ-like empirical refit from a short activation capture (PGKQ-KV$'$). Hadamard is disabled for V, where it helped fidelity in only 2/6 projection layers. Figure~\ref{fig:gate-m1-pareto} budgets this extension and plots mean LCDF $\Delta$ over 30 TinyStories sequences. Being a different, single-dataset measurement, Figure~\ref{fig:gate-m1-pareto} is not comparable to Table~\ref{tab:lcdf}. No both-lane configuration passes GATE-M1 ($\le 3.5$ bpe at $\Delta \le 0.05$ without being Pareto-dominated by KIVI); accordingly, we do not claim PGKQ-KV compression and confine the paper's compression claim to PGKQ's K-only setting. Every configuration differs significantly from KIVI (Holm-corrected p-values all $<0.001$; $p<0.001$ in all configurations in the paired Wilcoxon signed-rank test).

Repeated on 30 SQuAD-v2 sequences, every configuration misses the pre-registered $\Delta \le 0.05$ bar at ctx$=512$: five are Pareto-dominated by KIVI and the sixth has a smaller budget but a larger $\Delta$ (i.e., every configuration loses on at least one axis, confirming that the TinyStories outcome in Figure~\ref{fig:gate-m1-pareto} is not an artifact of that dataset). PGKQ-KV $\Delta$'s range from 0.738 (k4v4) to 3.394 (k2v2) across 2.74--4.74 bpe (k3v2: 3.195; k4v2: 3.146). The empirical-refit variant $'$ is uniformly closer to FP16 (e.g., k3v3 $\Delta$ 0.979 to 0.536) yet still far from competitive.

\subsection{KV-cache backend comparison}
\label{sec:kv-comparison}
PGKQ's throughput of 99.16 tok/s is about 43\% of FP16 (232.16 tok/s). We believe this gap is implementation-driven rather than intrinsic: unlike KIVI, PGKQ currently takes an unfused PyTorch path through the model's \texttt{estimate\_attention\_logits} hook (Hadamard rotation on $Q$, polar reconstruction, and QJL correction), whereas KIVI's scalar quantize--dequantize remains inside cuBLAS; we did not isolate fusion as the sole cause. PGKQ-KV (KV quantization as well) reaches 75.47 tok/s ($\sim$33\% of FP16) with a KV-byte budget of 184\,320 that fails GATE-M1; it is dominated by KIVI on cost and by K-only PGKQ on speed.

Table~\ref{tab:kv-comparison-full} reports a short-context comparison across eight KV-cache backends and six datasets on a single NVIDIA RTX~5090 GPU. Following the protocol, each benchmarked model runs a $1\times 64$ prefill (batch size 1, 64 prompt tokens) and a 64-token decode; this 128-token setting explains why FP16's TinyStories perplexity here (5.461) is higher than its Table~\ref{tab:base-training} mean under the full-context protocol (5.117)---the difference comes from the evaluation protocol, not from any change to the TinyStories checkpoint. All measurements are averaged over 10 random seeds (each seed is further repeated 30 times); numbers show the mean and range over datasets. The perplexity is the same for all eight KV-caches in a given dataset; differences across datasets reflect domain shift in the underlying (unchanged) model, not KV-cache contents. The first backend, \texttt{fw-default}, uses the model's native KV-cache path: relative to FP16 it differs only in avoiding our FP16 wrapper's extra indirection, so its KV-byte figure is n/a because the native path exposes no accounting hook. We omit the energy column because the GPU's host machine does not support energy measurement.

\begin{table}[t]\centering\footnotesize\setlength{\tabcolsep}{1.4pt}
\caption{RTX~5090 backend comparison}
\label{tab:kv-comparison-full}
\begin{tabular}{l c c c}
\toprule
Backend & TPS mean [range] & p99 ms mean [range] & KV bytes \\
\midrule
fw-default, native & 265.94 [265.49, 266.48] & 3.91 [3.87, 3.96] & N/A \\
\textbf{FP16} & 232.16 [231.97, 232.45] & 4.46 [4.46, 4.47] & 786\,432 \\
KIVI & 197.80 [196.87, 198.86] & 5.23 [5.17, 5.26] & 147\,456 \\
KVQuant-style & 187.14 [186.89, 187.43] & 5.54 [5.51, 5.56] & 172\,032 \\
QuaRot & 94.97 [94.89, 95.01] & 10.81 [10.77, 10.84] & 196\,608 \\
SnapKV & 219.33 [218.41, 220.03] & 4.73 [4.70, 4.76] & 786\,432 \\
PGKQ-KV, Beta & 75.47 [75.28, 75.68] & 13.56 [13.51, 13.60] & 184\,320 \\
\textbf{PGKQ, K-only} & 99.16 [99.01, 99.43] & 10.34 [10.32, 10.37] & 485\,376 \\
\bottomrule
\end{tabular}
\par\vspace{2pt}\parbox{\columnwidth}{\footnotesize Six datasets, ten seeds, 30 runs per seed. KV bytes are measured directly and are dataset-invariant; TPS and p99 give the mean [range] over datasets.}
\end{table}

\subsection{LCDF results}
\label{sec:lcdf-results}
On in-domain TinyStories (ctx$=512$; Table~\ref{tab:lcdf} gives six-dataset means), $\Delta$LCDF is $+0.002$ (KIVI), $+0.027$ (KVQuant), $+0.008$ (QuaRot), $<10^{-4}$ (SnapKV), 0.281 (PGKQ K), 0.546 (k3v3), and 0.083 (k4v4). ClinTemplate-synth and IoT-NLP-synth ran on CPU after a loader fix; GPU re-runs agree in all 14 signs.

\begin{table}[t]
\centering\footnotesize\setlength{\tabcolsep}{1.4pt}
\caption{Mean LCDF $\Delta$ against FP16 at ctx$=512$}
\label{tab:lcdf}
\begin{tabular}{l c c c c}
\toprule
Backend & Mean $\Delta$ [range] & Sig. & $\delta$ & 124M, sig. \\
\midrule
KIVI & 0.0053 [$-$0.0032,0.0239] & 3/6 & [$-$0.05,0.60] & $-$0.001; 3/6 \\
KVQuant-style & 0.0254 [0.0084,0.0478] & 6/6 & [0.03,1.00] & $-$0.006; 5/6 \\
QuaRot & 0.0117 [$-$0.0128,0.0333] & 6/6 & [$-$0.38,1.00] & $-$0.009; 5/6 \\
SnapKV & $|\Delta| < 10^{-4}$ & 2/6 & [$-$0.03,0.33] & $<\!10^{-4}$; 2/6 \\
PGKQ K & 0.1908 [0.0795,0.2806] & 6/6 & [0.17,1.00] & 0.037; 3/6 \\
PGKQ-KV k3v3 & 0.3190 [0.1982,0.5463] & 6/6 & [0.38,1.00] & 0.112; 6/6 \\
PGKQ-KV k4v4 & 0.1082 [0.0427,0.1937] & 6/6 & [0.10,1.00] & 0.063; 3/6 \\
\bottomrule
\end{tabular}
\par\vspace{2pt}\parbox{\columnwidth}{\footnotesize 30 decodes (ten seeds) on six datasets; brackets give the dataset range. Sig.: datasets with Holm-corrected $p<0.05$. $\delta$: Cliff's delta range. PGKQ rows use an empirical refit, unlike the Beta-prior sweep in Section~\ref{sec:pgkq-kv}. 124M: same protocol on the undertrained GPT-2 scale check.}
\end{table}

\begin{figure}[t]
\centering
\includegraphics[width=\columnwidth]{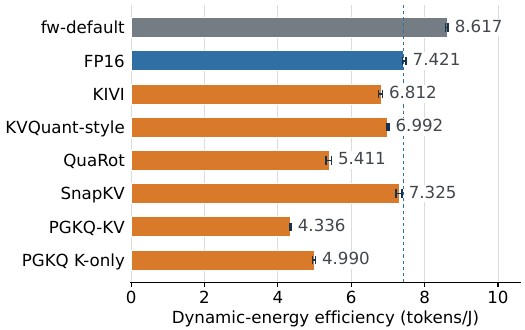}
\caption{Raspberry Pi~5 energy efficiency over ten seeds, 30 runs per seed, and six datasets. Bars show mean tokens per joule; whiskers give dataset ranges, and the dashed line marks FP16. Every compressed-cache backend falls below FP16 under our ports.}
\label{fig:pi5-energy}
\end{figure}

\begin{table}[t]
\centering\footnotesize\setlength{\tabcolsep}{1.4pt}
\caption{PSPCA versus one-shot pruning at matched sparsity}
\label{tab:pspca}
\begin{tabular}{l c c c c c c}
\toprule
 & & \multicolumn{5}{c}{TinyStories ppl} \\
\cmidrule(lr){3-7}
Target & Recov. & PSPCA & W & M & SG rand\,/\,real & Rand. \\
\midrule
0.30 & $4.1 {\times} 10^{-4}$ & 5.60$\pm$0.11 & 5.65$\pm$0.06 & \textbf{5.39} & 5.49\,/\,5.36 & 56.5 \\
0.50 & $4.2 {\times} 10^{-3}$ & 7.08$\pm$0.17 & \textbf{7.06$\pm$0.13} & 7.19 & 7.70\,/\,5.84 & $5.0{\times}10^{3}$ \\
0.70 & $2.1 {\times} 10^{-2}$ & 19.77$\pm$0.53 & 24.22$\pm$1.06 & \textbf{16.90} & 26.51\,/\,10.62 & $1.4{\times}10^{8}$ \\
\bottomrule
\end{tabular}
\par\vspace{2pt}\parbox{\columnwidth}{\footnotesize Eight-round PSPCA vs.\ one-shot Wanda (W), magnitude (M), and SparseGPT (SG) at matched per-linear sparsity. TinyStories ppl is mean$\pm$sd over five seeds; dense is 5.461; M is data-free and SG deterministic (random\,/\,real-text calibration). Rand.: three-seed random-mask floor. Recov.: final anchor-recovery loss. Bold: best matched mean.}
\end{table}

\subsection{PSPCA pruning}
\label{sec:pspca-results}
\textit{Magnitude pruning:} One-shot magnitude pruning doesn't take gradient steps and achieves lower perplexity at 30 and 70\% sparsity than all other methods we've tried except real-text-calibrated SparseGPT. At 30\% it achieves a perplexity of 5.39 (vs.\ PSPCA's $5.60\pm0.11$) lower than the perplexity of the unpruned model (5.461), and at 70\% its perplexity of 16.90 is lower than any seed's perplexity ($\ge 19.0$, see Table~\ref{tab:pspca}), an advantage of $\sim$5 seed std.\ dev.

\textit{PSPCA:} Our negative PSPCA verdict holds over five seeds. At 50\%, the reported advantage does not exceed one standard deviation of PSPCA seed-to-seed perplexity (7.08 PSPCA, 7.19 mag.\ prune) and is considered indistinguishable. It does not significantly outperform Wanda at 30 and 50\%, and at 70\% the results are significantly better than Wanda with non-overlapping seed-to-seed ranges. We replicated the published single-seed perplexities using our evaluation pipeline and additionally replicated the results of the one-shot pruning algorithms bit-for-bit and PSPCA to within 0.4. All methods are orders of magnitude better than the random mask baseline.

\textit{Sanity check SparseGPT results on TinyStories with real calibration data:} SparseGPT calibrated with random token ids (same as Wanda and PSPCA) performs the worst of the methods we've tried at 50 and 70\% sparsity, however, in a sanity check where we calibrate with real TinyStories sentences SparseGPT outperforms all methods at all sparsity levels. This suggests that the quality of the calibration data could be more important than the pruning algorithm, however, we have not tested this claim with a second architecture.

\subsection{Energy and latency on Raspberry Pi 5}
\label{sec:pi5-results}
Figure~\ref{fig:pi5-energy}. We report power using INA219 draw minus the measured idle draw of 2.51\,W. Our cooled Pi~5 does not throttle. As dynamic draw is similar at 4.72--4.82\,W, energy efficiency trends with throughput. The native cache path is more efficient (8.617 versus 7.421 tok/J) than FP16 and all compressed settings trail FP16 \cite{ardakani2025llmpi}. PGKQ reaches 67.9\% of FP16 throughput, compared to 42.7\% on GPU.

\section{Limitations and threats to validity}
\label{sec:limitations}
\textit{Limitations:} Most results use the 8.39\,M-parameter TinyStories-trained model; one check at 124\,M changes some rankings within its model family, and we do not study models with 1B parameters or more. Our experiments are modest in compute (8.9 RTX~5090 GPU-hours in total), which may miss improvements that require more resources. Cache sizes are relatively small (up to 3.1\,MB on the main evaluation platform, 18.9\,MB in the scaling study), so gains come from improving accuracy within feasible budgets rather than making previously infeasible sizes feasible.

Results are strongest on TinyStories; the other datasets serve as distribution-shift tests and two are synthetic. We do not evaluate code, multilingual, instruction-following, or retrieval-heavy settings. LCDF is evaluated with next-token log-loss rather than downstream-task or broader semantic-quality metrics. PSPCA gets recovery opportunities that one-shot baselines lack; its comparatively weak results are therefore not due to such an imbalance. The destructive-cache protocol is a worst-case robustness/sensitivity measure, not necessarily reflective of user-perceived importance. A mixed-effects analysis across datasets, prompts, and seeds is left to future work. Our ablations and controls cover two datasets. QuaRot and SnapKV baselines are our own implementations validated via the destructive-cache task; differences from other implementations are possible. KIVI agrees with the reference implementation up to logits, but our KVQuant-style K lane diverges from the reference. Some experiments (the PGKQ rows for Pi~5, cross-device verifiability, the headline benchmarks, and the CPU experiments) use a corrected angle-bitwidth configuration; other Pi~5 experiments use the original parameter sweep, and the PGKQ investigation at 124\,M is a separate campaign. The corrected rows are not aggregated with other results.

\section{Conclusion}
\label{sec:conclusion}
The EdgeLLM-Nano study suggests that results for KV-cache compression for edge LLM inference may not generalize well and can be benchmark-dependent. Neither of our experimental methods achieved their goal according to our benchmark that stresses KV-cache: LCDF. PGKQ achieves 3.75 bits/K element but an RTN baseline at a comparable bit-width degrades less; KIVI outperforms any of the two-lane configurations we considered; PSPCA with 1\,600 recovery steps does not reliably outperform one-shot magnitude pruning. On the target device, the Raspberry Pi~5, the energy-efficiency of compressed methods is less than FP16 in tokens/J: the software overheads hurt throughput more than they save power. The suite of statistical tests in LCDF, including destructive-cache sanity check, seed-level paired test, and effect-size estimation, will be released, in the hope that the motivating analysis and audit process can encourage broader adoption of LCDF as part of the methodology of KV-cache research.

\section*{Acknowledgment}
Claude by Anthropic was used for the implementation. All experiments and ideas were done by the authors. We take full responsibility for this work.

\clearpage
\balance
\bibliographystyle{IEEEtran}
\bibliography{references}

\end{document}